# New Results for Domineering from Combinatorial Game Theory Endgame Databases

Jos W.H.M. Uiterwijk[a,*], Michael Barton[a]

[a]*Department of Knowledge Engineering (DKE), Maastricht University P.O. Box 616, 6200 MD Maastricht, The Netherlands*

**Abstract**

We have constructed endgame databases for all single-component positions up to 15 squares for Domineering. These databases have been filled with exact Combinatorial Game Theory (CGT) values in canonical form. We give an overview of several interesting types and frequencies of CGT values occurring in the databases. The most important findings are as follows.

First, as an extension of Conway's [8] famous Bridge Splitting Theorem for Domineering, we state and prove another theorem, dubbed the Bridge Destroying Theorem for Domineering. Together these two theorems prove very powerful in determining the CGT values of large single-component positions as the sum of the values of smaller fragments, but also to *compose* larger single-component positions with specified values from smaller fragments. Using the theorems, we then prove that for any dyadic rational number there exist single-component Domineering positions with that value.

Second, we investigate Domineering positions with infinitesimal CGT values, in particular ups and downs, tinies and minies, and nimbers. In the databases we find many positions with single or double up and down values, but no ups and downs with higher multitudes. However, we prove that such single-component ups and downs easily can be constructed, in particular that for any multiple up $n{\cdot}\uparrow$ or down $n{\cdot}\downarrow$ there are Domineering positions of size $1+5n$ with these values. Further, we find Domineering positions with 11 different tinies and minies values. For each we give an example. Next, for nimbers we find many Domineering positions with values up to *3. This is surprising, since Drummond-Cole [10] suspected that no *2 and *3 positions in standard Domineering would exist. We show and characterize many *2 and *3 positions. Finally, we give some Domineering positions with values interesting for other reasons.

Third, we have investigated the temperature of all positions in our databases. There appears to be exactly one position with temperature 2 (as already found before) and no positions with temperature larger than 2. This supports Berlekamp's conjecture that 2 is the highest possible temperature in Domineering.

*Corresponding author.
*Email address:* `uiterwijk@maastrichtuniversity.nl` (J.W.H.M. Uiterwijk)

## 1. Introduction

Domineering is a two-player perfect-information game invented by Göran Andersson around 1973. It was popularized to the general public in an article by Martin Gardner [13]. It can be played on any subset of a square lattice, though mostly it is restricted to rectangular $m \times n$ boards, where $m$ denotes the number of rows and $n$ the number of columns. The version introduced by Andersson and Gardner was the $8 \times 8$ board.

Play consists of the two players alternately placing a $1 \times 2$ piece (domino) on the board, where the first player may place the piece only in a vertical alignment, the second player only horizontally. Dominoes may not overlap. The first player being unable to move loses the game, his opponent (who made the last move) being declared the winner. Since the board is gradually filled, i.e., Domineering is a converging game, the game always ends, and ties are impossible. With these rules the game belongs to the category of *partizan combinatorial games*, for which a whole theory (the Combinatorial Game Theory or CGT in short) has been developed, especially by Conway [8] and Berlekamp et al. in their famous book set *Winning Ways* [4]. In CGT the first player conventionally is called Left, the second player Right, though in our case we will also use the more convenient indications of Vertical and Horizontal, for the first and second player, respectively.

Among combinatorial game theorists Domineering received quite some attention, but this was limited to rather small or irregular boards [1, 3, 4, 8, 15, 26]. Larger (rectangular) boards were investigated using $\alpha$-$\beta$ search [16], leading to solving all boards up to the standard $8 \times 8$ board [6], later extended to the $9 \times 9$ board [25, 14], and finally extended to larger boards up to $10 \times 10$ [7].

In order to combine the best of two worlds, namely the detailed analysis and values of the CGT viewpoint and the power of uninformed search, we decided to incorporate CGT knowledge into an $\alpha$-$\beta$ solver for Domineering [2]. As part of that we built (endgame) databases with exact combinatorial values for all single-component Domineering positions up to 15 squares.

A statistical analysis of these CGT databases revealed interesting findings, which we report in this article. Section 2 gives relevant background. First some related research will be described. We then give some basic facts from CGT, to be used further. The section ends with a short overview how the databases were constructed. Next, in Section 3, we give overall statistics on the contents of the databases and then report on special types of CGT values encountered. In particular, we concentrate on numbers and infinitesimals (ups and downs, tinies and minies, and nimbers). Also some other values of special interest are given. In Section 4 we report our findings on temperatures encountered in Domineering positions. In Section 5 we conclude and give some directions for future research.

## 2. Background

In this section we first give an overview of previous research done on CGT values for Domineering, and related research, mostly concerning the game of Amazons (Subsection 2.1). Then, we give an introduction to the CGT, especially concerning several types of CGT values (Subsection 2.2), and show Conway's [8] Bridge Splitting Theorem for Domineering (Subsection 2.3). Finally, we describe how the endgame databases for Domineering, used throughout this research, have been constructed (Subsection 2.4).

### *2.1. Related Research*

Some related work on Domineering has been done by Conway [8] and by Berlekamp, Conway, and Guy [4], where they provided CGT values for all Domineering positions up to 7 squares, and for some positions up to 12 squares. Further they provided values for several sequences of Domineering positions. Berlekamp [3] determined CGT values for many classes of $2 \times n$ and $3 \times n$ Domineering games, based on the idea of generalized overheating and heating these games. Wolfe [26] and Kim [15] analyzed 1-dimensional Domineering positions with corners and kinks (so-called snakes, see also Subsection 3.2). Shankar and Sridharan [18] investigated many Domineering positions for their temperatures. Another way to solve Domineering boards has been reported by Lachmann *et al.* [17], who used translational symmetry rules to establish game-theoretic values for boards based on the values for smaller subboards. This work recently was updated and corrected by Drummond-Cole [11]. No endgame databases were used in the research reported in all these publications.

Most other related work has been done in the area of the game Amazons. Tegos [23] implemented CGT endgame databases for Amazons. These contained just thermograph information, not exact CGT values, and therefore only could be used for (heuristic) move-ordering purposes. Snatzke [20] built CGT endgame databases for a very restricted version of Amazons, namely for positions fitting within a $2 \times 11$ board with exactly 1 queen per player. This was extended in [21] with new results for some small databases of other shapes, with 1 to 4 queens. Recently, Song [22] implemented endgame databases in an Amazons solver. Again, the databases did not contain exact CGT values, but heuristic (thermograph) information useful for narrowing the bounds in the solving process. With his program $5 \times 6$ Amazons has been solved.

As far as we know there is no other literature reporting the use of endgame databases with precise CGT values in an $\alpha$-$\beta$ solver.

### *2.2. Introduction to the Combinatorial Game Theory*

Here we first give a general introduction to the CGT, in so far as applicable to Domineering. In particular we report on definitions and meaning of CGT values as reported in Section 3. For more detailed information on CGT we refer to many standard works in this area, in particular [8, 4, 1].

In CGT applied to Domineering it is common to indicate the two players by Left (Vertical) and Right (Horizontal). A game is then described by its left (vertical) and right (horizontal) options, where options are again games. A left option describes a legal move Left can make by giving the game position after performing the move, whereas a right option describes a legal move Right can make. To better understand this recursive definition, it is easiest to start by analyzing the simplest of games. A game of Domineering consisting of no or merely one square (see Figure 1 left) has no legal move for either player, i.e., no left options and no right options, and is called zero or 0. In short, since games are represented by their left and right options, this game is described by $G = \{|\} = 0$. The second and third game in Figure 1 show the two next simplest Domineering games, 1 and $-1$, that each have exactly one legal move for Left or Right, respectively. Using the previously shown game notation of left and right options, they are described by $1 = \{\{|\}\ |\} = \{0\ |\}$ and $-1 = \{|\ \{|\}\} = \{|\ 0\}$. By expanding the game board to larger and more arbitrary shapes, very complex representations can be obtained. Only a small portion of the larger games have easy values such as 0, 1 or $-1$. For example, the fourth game in Figure 1 shows the slightly more complicated value of $G = \{\{\{|\}\ |\} \parallel \{|\ \{|\}\}\} = \{\{0\ |\} \parallel \{|\ 0\}\} = \{1 \mid -1\} = \pm 1$. It is generally true that games with positive values pose an advantage for Left, and games with negative values pose an advantage for Right.

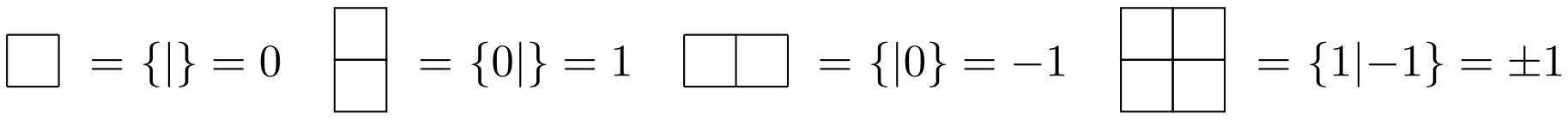

Figure 1: From left to right, the simplest Domineering positions with values 0, 1, $-1$, and $\pm 1$.

### *2.2.1. Numbers*

The games 0, 1, and $-1$ have values that are called numbers. Numbers may be positive and negative integers, but can also be fractions. In games with finite game trees specifically, all numbers are *dyadic rationals*, i.e., rational numbers whose denominators are a power of two. Numbers are represented by the options for both Left and Right: $G = \left\{G^L \;\middle|\; G^R\right\}$, where $G^L$ and $G^R$ are either $\{\} = \emptyset$ or sets of other numbers. For all numbers with two non-empty options it holds that $G^L < G^R$. Generally, the sign of a number indicates which player wins the game, while the value 0 indicates a win for the player who made the last move, i.e., the previous player. Numbers and combinations of multiple numbers thus have a predetermined outcome.

### *2.2.2. Infinitesimals*

Infinitesimal games have values that are very close to zero and are less than every number with the same sign. Being infinitesimally close to zero, these games can never overpower any game with non-zero number value under optimal play. There are several types of infinitesimal values, and this subsection introduces three of them: ups/downs, tinies/minies, and nimbers.

*Ups/Downs*

The two games up and down are opposites and defined as $\uparrow = \{0 \mid *\}$ and $\downarrow = \{* \mid 0\}$, respectively, where $*$ is a nimber (see below). Up is a positive infinitesimal, and down is its opposite, thus negative, i.e., $\downarrow = -\uparrow$. These positions arise frequently in combinatorial games and pose an avertable threat to one of the players. An example of an up position and its left and right options is given in Figure 2. In this game, Left can best move to 0. If Right gets to play instead, he can only move to a $*$ position. In the game down, Left's and Right's roles are reversed.

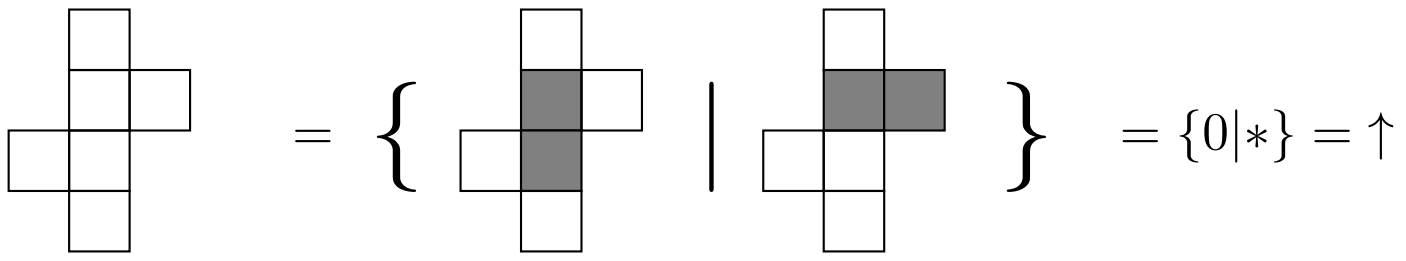

Figure 2: A Domineering position with value $\uparrow$ and its optimal Left and Right moves.

*Tinies/Minies*

Tiny and miny games are a group of games that, like encountered above, hold out the prospect of a favorable number for one player that the other player can forestall. Tiny-$x$ is positive and is defined as $⧾_x = \{0 \mid \{0 \mid -x\}\}$ with $x \geq 0$. Miny-$x$ is its opposite, thus negative, and defined as $⧿_x = \{\{x \mid 0\} \mid 0\}$, where again $x \geq 0$. From the definitions we see that up and down are in fact just other names for $⧾_0 = \{0 \mid \{0 \mid 0\}\} = \{0 \mid *\} = \uparrow$ and $⧿_0 = \{\{0 \mid 0\} \mid 0\} = \{* \mid 0\} = \downarrow$ that arose due to their special characteristics compared with $⧾_x$ and $⧿_x$ for $x > 0$. Figure 3 shows a position with value $⧾_2$ for Domineering.

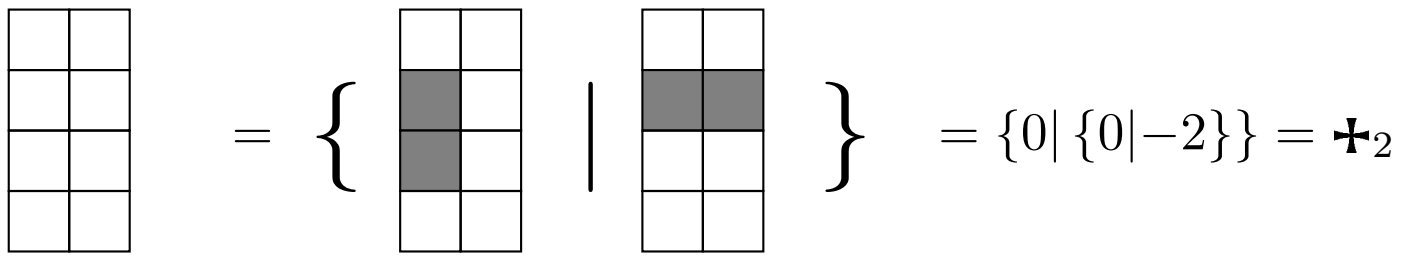

Figure 3: A Domineering position with value $⧾_2$ and its optimal Left and Right moves.

*Nimbers*

Nimbers commonly occur in impartial games. Their main characteristics are that their left and right options are identical, following a certain schema, and that they are their own negatives. Nimbers are defined by $*(n+1) = \{*0, \ldots, *n | *0, \ldots, *n\}$. Here, $*0 = 0$ is the only nimber that also is a number. The next nimber, $*1$ (or just $*$ in short) $= \{0|0\}$. While there may theoretically occur nimbers in the game of Domineering, which is not an impartial game, the 0 and $*$ are the only frequent representatives of this group according to Drummond-Cole [10]. Figure 4 shows a position with value $*$ in Domineering. Clearly, the first player to move in this game wins, which is why the outcome is not predetermined and why its value is truly confused with zero. This introduces an interesting property of nimbers: every nimber is its own negative, i.e., $*n = -*n$. As a consequence, when two identical nimbers occur, they cancel each other: $*n + *n = 0$, where $n$ is any positive integer.

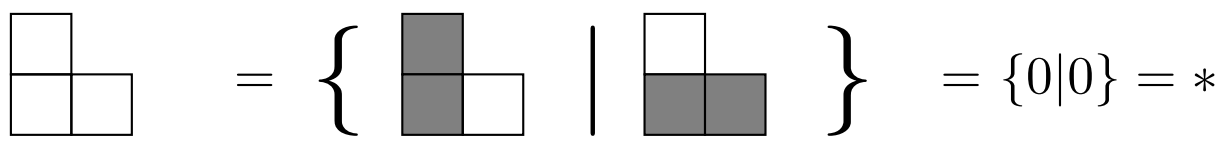

Figure 4: A Domineering position with value $*$ and its Left and Right moves.

*Comparison of Scale*

All three types of games that are described above are infinitesimal. That means, their values are closer to 0 than any non-zero number, but still distinct. Even though their values are very small, it is possible to compare infinitesimals to another, and to position them around 0. This is what Figure 5 illustrates.

The very smallest of games are tinies. As is described above, tinies take on a number $x \geq 0$ that defines their value. The larger $x$ is, the smaller the value of $⧾_x$ becomes. More specifically, for two numbers $y > x \geq 0$, $⧾_y$ is infinitesimal with respect to $⧾_x$. Consequently, there exists no finite multiple of $⧾_y$ that exceeds $⧾_x$ in value, i.e. $\nexists n : n \cdot ⧾_y \geq ⧾_x$ for any positive integer $n$. Following this rule, $⧾_0 = \uparrow$ is the largest tiny game. This leaves the only unusual infinitesimal, the star. The value of $*$ is incomparable to or confused with 0 and with $\uparrow$, denoted by $0 \parallel * \parallel \uparrow$, but is smaller than $2 \cdot \uparrow = \Uparrow$.

On the other side of 0, negative infinitesimals scale exactly as their positive counterparts do: $2 \cdot \downarrow = \Downarrow < * \parallel \downarrow \leq ⧿_x < ⧿_y < 0$, where $x$ and $y$ are numbers such that $y > x \geq 0$.

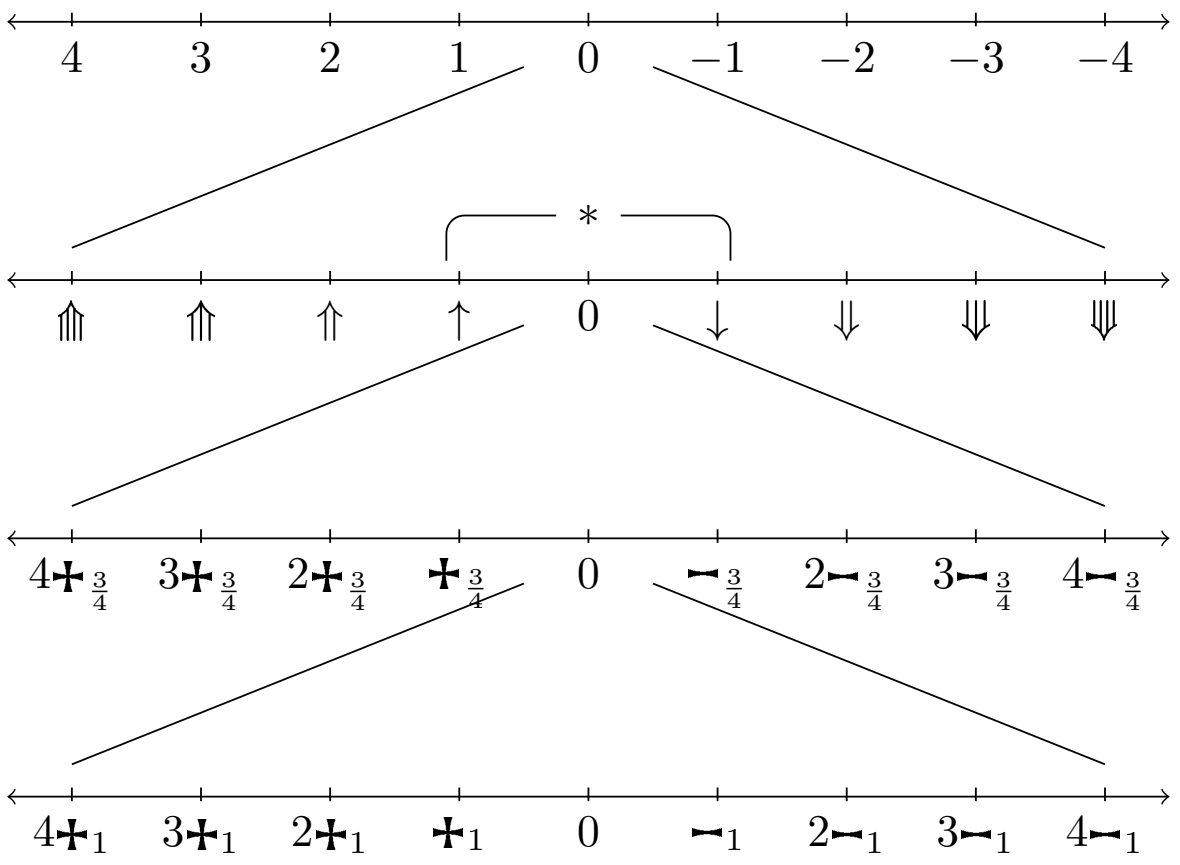

Figure 5: A comparison of scale of infinitesimals star, ups/downs and tinies/minies with respect to simple numbers as taken from [1]. Note: the axes directions are exchanged to emphasize the player, Left or Right, that attains an advantage.

*2.3. The Bridge Splitting Theorem*

Though the preceding theory is applicable to any combinatorial game, Conway formulated a beautiful theorem that is specifically applicable to Domineering [8]. Because of its importance, we repeat the theorem and its proof. We denote it as the *Bridge Splitting Theorem*, to contrast it with another theorem to be formulated later.

**Theorem 1** (Bridge Splitting Theorem for Domineering)**.**
*If for some game $G\Box$ its value is equal to that of the game $G$ alone, then the value of $G\Box H$ is the sum of the values of $G$ and $\Box H$, provided that $G$ and $H$ have no common edges.*

*Proof.*

$$G\Box H \leq G + \Box H = G\Box + \Box H \leq G\Box H$$

The first inequality is justified, since splitting a horizontal line can only favor Vertical. The equality is the condition of the theorem. The second inequality is justified since joining two horizontally adjacent squares also can only favor Vertical. □

Following Conway [8] we denote such bridging square as *explosive.* Note that this theorem concerns two fragments connected horizontally. Of course the theorem is equally valid when the two fragments are vertically connected. Important is that the two fragments are linearly connected via an explosive bridge, i.e., that the bridge has two **opposite** edges in common with the two fragments.

*2.4. Construction of the CGT Endgame Databases*

In order to use CGT values for Domineering games we have built all Domineering endgame databases for positions of sizes 1–15. Positions that exhibit mirror symmetry (left/right and/or top/bottom) are unified into a single entry. No 90° rotation symmetry, i.e., switching the players, is applied at present. The CGT values for the Domineering games are calculated with the CGSUITE software tool [19]. These values are in so-called canonical form, which make them a unique representation of the values.

## 3. Values in Domineering

In this section we give an overview of the content of our databases. First we give a statistical overview of our databases (see Subsection 3.1). We then concentrate on “pure” values encountered, i.e., *numbers* (Subsection 3.2), *ups* and *downs* (Subsection 3.3), *tinies* and *minies* (Subsection 3.4), and *nimbers* (Subsection 3.5). We finally give some interesting other values in Subsection 3.6.

*3.1. Statistical Analysis of the Databases*

We have generated the endgame databases up to a maximum position size of 15 squares. Beyond that point, the number of entries quickly grows too large to fit in the average home computer's main memory. A survey of the completed databases shows that the number of entries approximately increases by a factor of between 3 and 4 with every additional square. At the final size of 15, the databases contain a total of about 9.3 million entries.

Table 1 gives an overview of how positions of a fixed size are distributed over the various types of CGT values. Generally, each column describes one distinct type of value or a subset of them. The first block accounts for the pure number games, with the first column listing the numbers of games that are zero, i.e., wins for the previous player, the second column accounting for all non-zero number games, and the third column summing all number games. The second block accounts for the pure infinitesimal games, with $n \cdot \uparrow / n \cdot \downarrow$ the numbers of non-zero multiples of ups and downs, $⧾_x / ⧿_x$ the numbers of tiny-$x$ and miny-$x$ games with $x > 0$, and the next column representing all nimber games except $*0 = 0$. The last column in this block sums the previous three columns plus the 0 games, i.e., gives the total number of pure infinitesimals for each size. To account for combinations of these basic values, the column Comb. counts all occurrences of the previous types and any combinations thereof. The last column lists the total numbers of games of a particular position size.

| Pos. size | Num. $= 0$ | Num. $\neq 0$ | Num. total | $n \cdot \uparrow$ $n \cdot \downarrow$ | $⧾_x$ $⧿_x$ | $*n$ $\neq *0$ | Inf. total | Comb. | Total |
|---|---|---|---|---|---|---|---|---|---|
| 2 | 0 | 2 | 2 | 0 | 0 | 0 | 0 | 2 | 2 |
| 3 | 0 | 2 | 2 | 0 | 0 | 1 | 1 | 3 | 3 |
| 4 | 0 | 4 | 4 | 0 | 0 | 2 | 2 | 6 | 9 |
| 5 | 5 | 4 | 9 | 0 | 0 | 1 | 6 | 16 | 21 |
| 6 | 10 | 16 | 26 | 2 | 0 | 0 | 12 | 38 | 68 |
| 7 | 13 | 48 | 61 | 4 | 0 | 13 | 30 | 106 | 208 |
| 8 | 16 | 194 | 210 | 6 | 4 | 46 | 72 | 320 | 730 |
| 9 | 116 | 386 | 502 | 2 | 6 | 104 | 228 | 950 | 2542 |
| 10 | 515 | 1262 | 1777 | 94 | 8 | 136 | 753 | 3125 | 9287 |
| 11 | 1061 | 3570 | 4631 | 336 | 0 | 462 | 1859 | 9867 | 34,053 |
| 12 | 2074 | 14,700 | 16,774 | 764 | 28 | 3618 | 6484 | 32,990 | 127,112 |
| 13 | 5012 | 45,018 | 50,030 | 1392 | 188 | 13,768 | 20,360 | 108,994 | 476,849 |
| 14 | 27,816 | 155,410 | 183,226 | 5018 | 820 | 24,002 | 57,656 | 367,330 | 1,803,636 |
| 15 | 135,539 | 437,718 | 573,257 | 23,752 | 1988 | 46,254 | 207,533 | 1,237,853 | 6,851,960 |
| Total | 172,177 | 658,334 | 830,511 | 31,370 | 3042 | 88,407 | 294,996 | 1,761,600 | 9,306,480 |

Table 1: Distribution of Domineering positions of a fixed size up to 15 over different types of CGT values.

Clearly, as the overall number of entries grows with increasing position size, the ratio of all simple value types gradually decreases. Even at the maximum size of 15, however, still approximately 8% of the entries are number games, 3% are pure infinitesimal games, and 18% are any combination of numbers and infinitesimals.

*3.2. Numbers*

We first repeat that the only numbers occurring in Domineering (and all finite combinatorial games) are *dyadic rationals*, i.e., those rational numbers whose denominator is a power of 2. An integer value $n$ occurs for a linear chain of $2n$ empty squares, vertical chains for positive integers (an $n$-move advantage for Vertical), horizontal chains for negative integers (an $n$-move advantage for Horizontal). 0 is the value of the empty game or the game consisting of a single empty square.

Before turning to other numbers, we first formulate, in analogy with the Bridge Splitting Theorem, another theorem, that can be seen as an extension of the Bridge Splitting Theorem.

**Theorem 2** (Bridge Destroying Theorem for Domineering)**.**
*If for some game $G\square$ its value is equal to that of the game $G$ alone, for some game $\overset{H}{\square}$ its value is equal to that of the game $H$ alone, for some game $\square I$ its value is equal to that of the game $I$ alone, and for some game $\underset{J}{\square}$ its value is equal to that of the game $J$ alone, then the value of $G\overset{H}{\underset{J}{\square}}I$ is the sum of the values of $G$, $H$, $I$, and $J$, provided that $G$, $H$, $I$, and $J$ have no common edges. Games $G$, $H$, $I$, and $J$ might be empty.*

*Proof.*

$$\begin{aligned}
G\overset{H}{\underset{J}{\square}}I \leq G + \overset{H}{\underset{J}{\square}}I \leq G + \overset{H}{\underset{J}{\square}} + I &= G + \overset{H}{\square} + I + J \\
&= G + H + I + J \\
&= G\square + H + I + J = G\square I + H + J \leq G\overset{H}{\square}I + J \leq G\overset{H}{\underset{J}{\square}}I
\end{aligned}$$

The first two inequalities are justified, since splitting a horizontal line can only favor Vertical, the first equality is an application of the Bridge Splitting Theorem for vertical connections, the next two equalities are just two conditions of the theorem, the fourth equality is an application of the Bridge Splitting Theorem for horizontal connections, and the last two inequalities are justified, since linking along a vertical line also can only favor Vertical. □

We note that this proof also includes the cases where any subset of $\{G, H, I, J\}$ are empty games, since the value of the empty square equals the value of an empty game. We further note that for two non-empty games this theorem is covered by the Bridge Splitting Theorem when the games are connected to opposite sides of the bridging square, but that this is not needed for the Bridge Destroying Theorem, i.e., a "corner" connection is also allowed.

The Bridge Destroying Theorem in combination with the Bridge Splitting Theorem proves very powerful in obtaining values for large loosely connected positions. It also proves useful in building complex single-component positions from smaller building blocks with as value the sum of the fragments.

As an example, suppose we want to build a single-component position with value 3/4. Using the building blocks depicted in Figure 6 with values 1/2 and 1/4, respectively, where extending the positions with the explosive (marked) squares does not change their values, we can easily build a position with value 3/4 by connecting both extended figures by joining the explosive squares (see Figure 7).

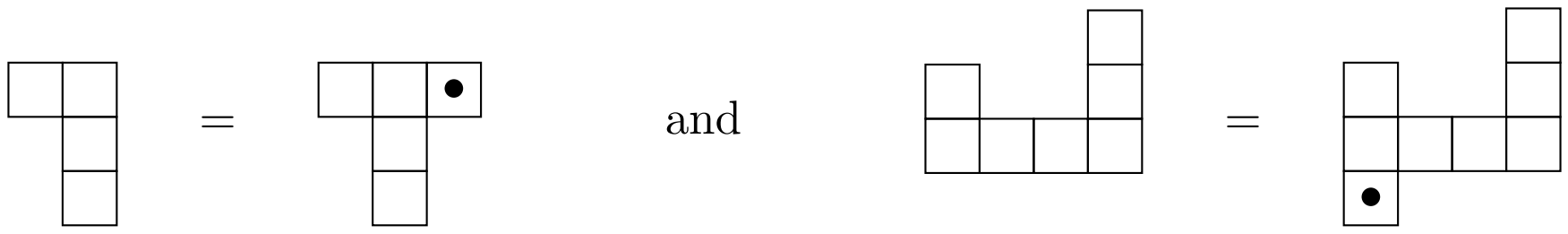

Figure 6: Building blocks with values 1/2 and 1/4.

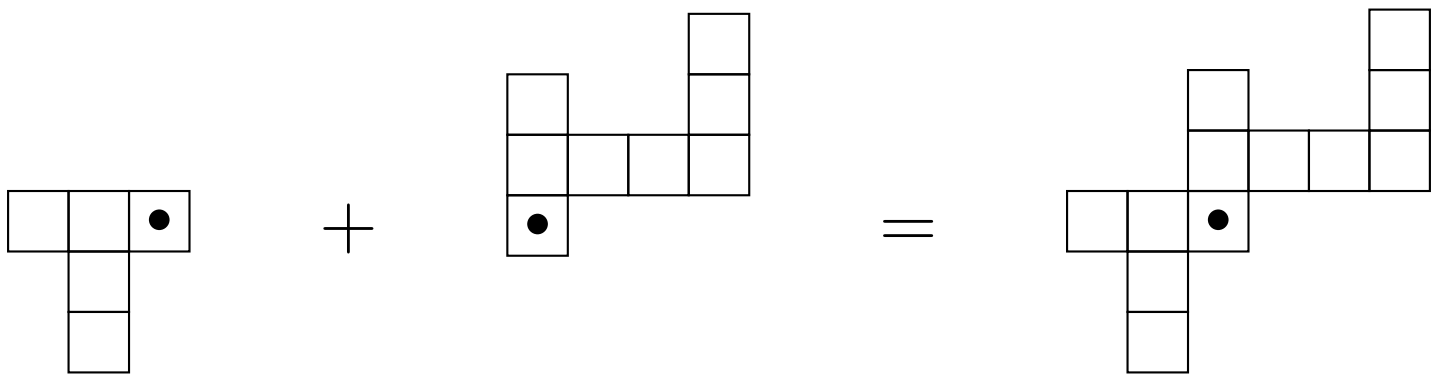

Figure 7: A position with value 3/4 as the sum of two fragments with values 1/2 and 1/4.

We are now ready to start generalizing to arbitrary dyadic rational numbers. In Figure 8 we provide some basic fractions. Using the fact that the marked squares of the four basic fractions in Figure 8 are explosive, we can build single-component positions with values 3/8, 5/8, and 7/8, by proper combinations, namely $3/8 = 1/8 + 1/4$, $5/8 = 1/8 + 1/2$, and $7/8 = 1/8 + 3/4$ (see Figure 9).

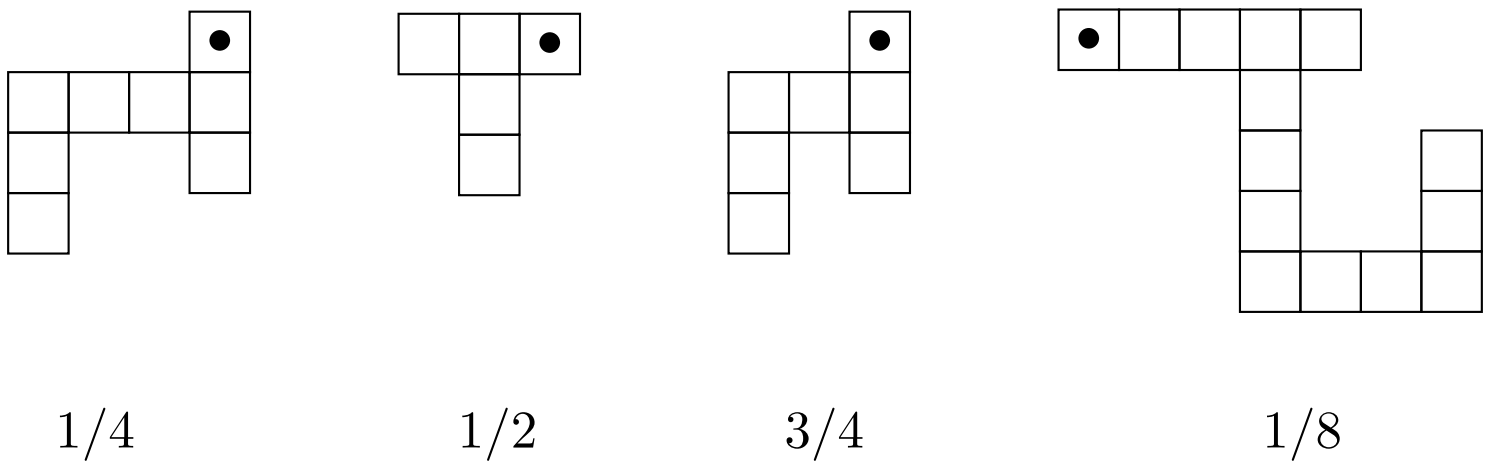

Figure 8: Some basic fractions. Explosive squares are indicated.

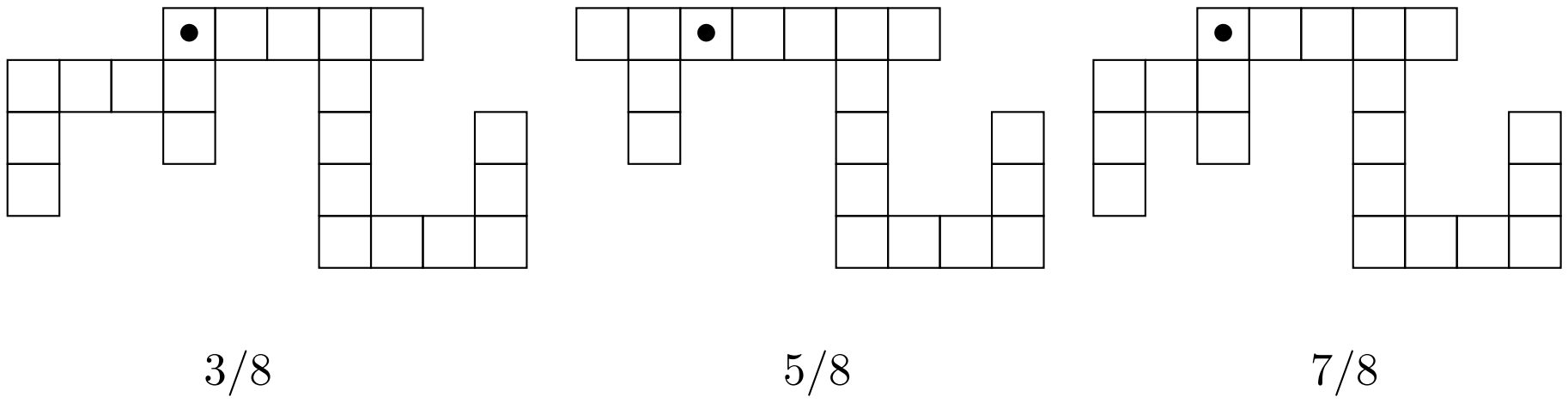

Figure 9: Some less basic fractions as combinations of basic fractions. Explosive squares are indicated.

To build fractions with larger denominators, we use the results by Kim [15] as reported by Guy [12], p. 477. Here, fragments are shown with fractions 1/16, 1/32, 1/64, etc., i.e., for all $1/2^k$, with $k \geq 4$. We repeat Guy's Figure 1 as our Figure 10, where we have flipped in every odd-numbered (from left to right) dashed box the right vertical tail (in order not to hamper our following construction), without affecting the values.

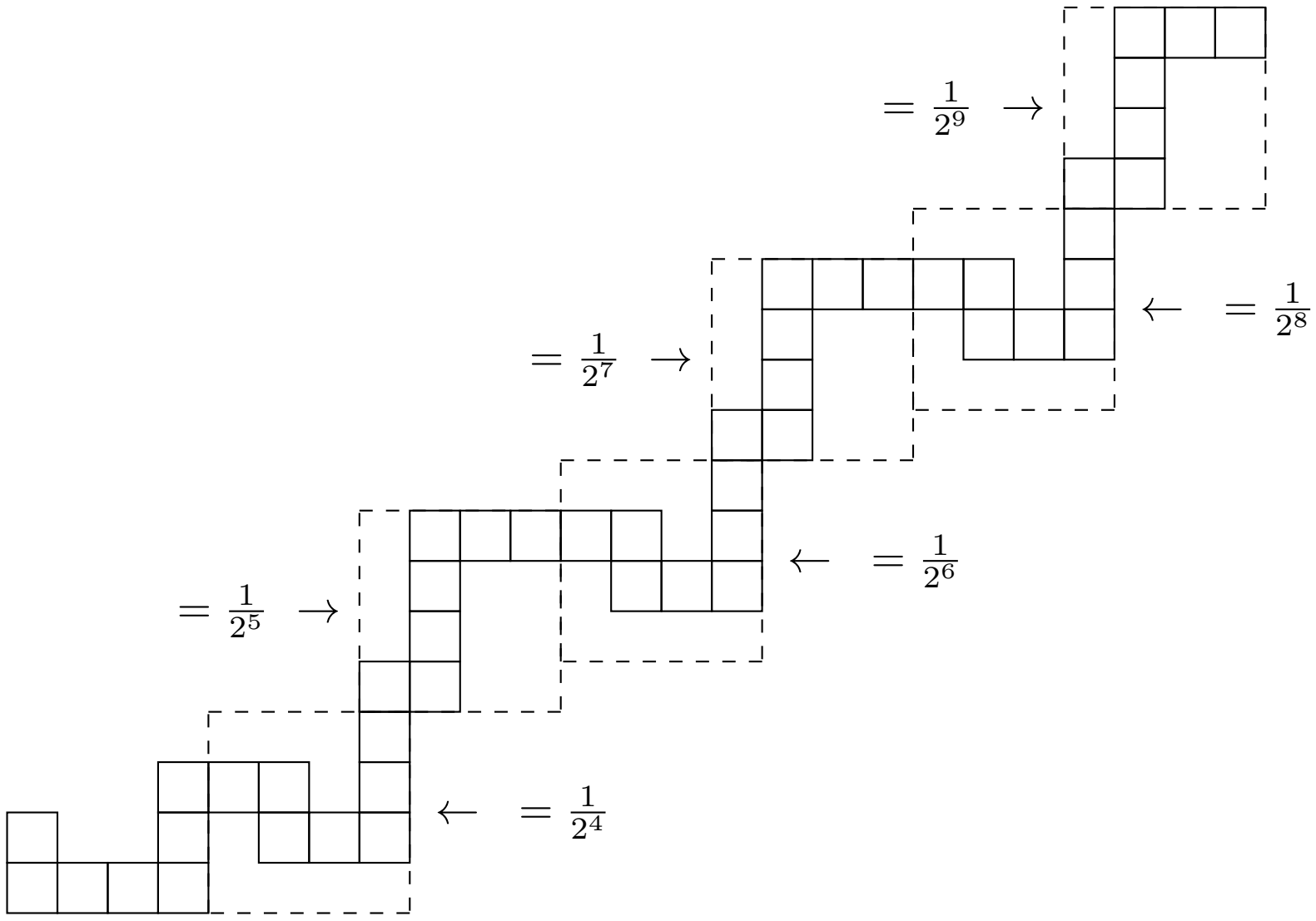

Figure 10: Yonghoan Kim's snakes of value $2^{-n}$ (adapted from [12], Figure 1, p. 477).

Starting with the fragment with value 1/16 from Figure 10, we note that we can add two explosive squares to the fragment as follows (see Figure 11).

As a result we have a single-component position with value 1/16, where we can use the two explosive squares as bridges with other equal fragments of value 1/16, as follows (see Figure 12).

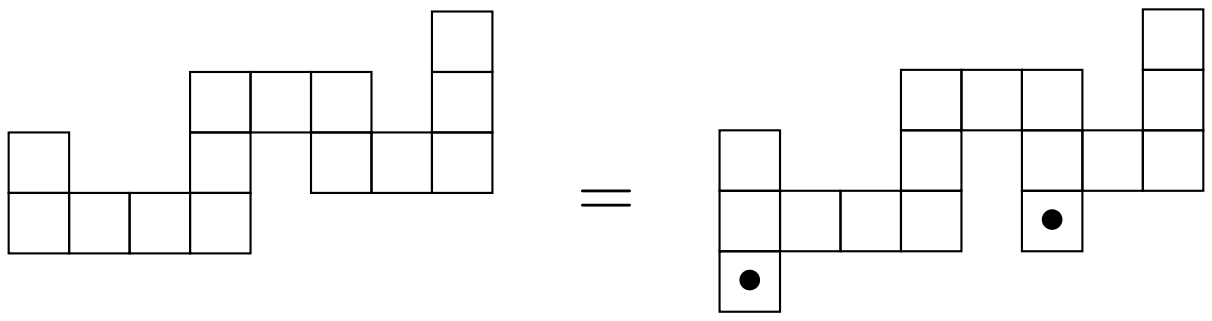

Figure 11: A fragment with value 1/16 and two explosive squares added.

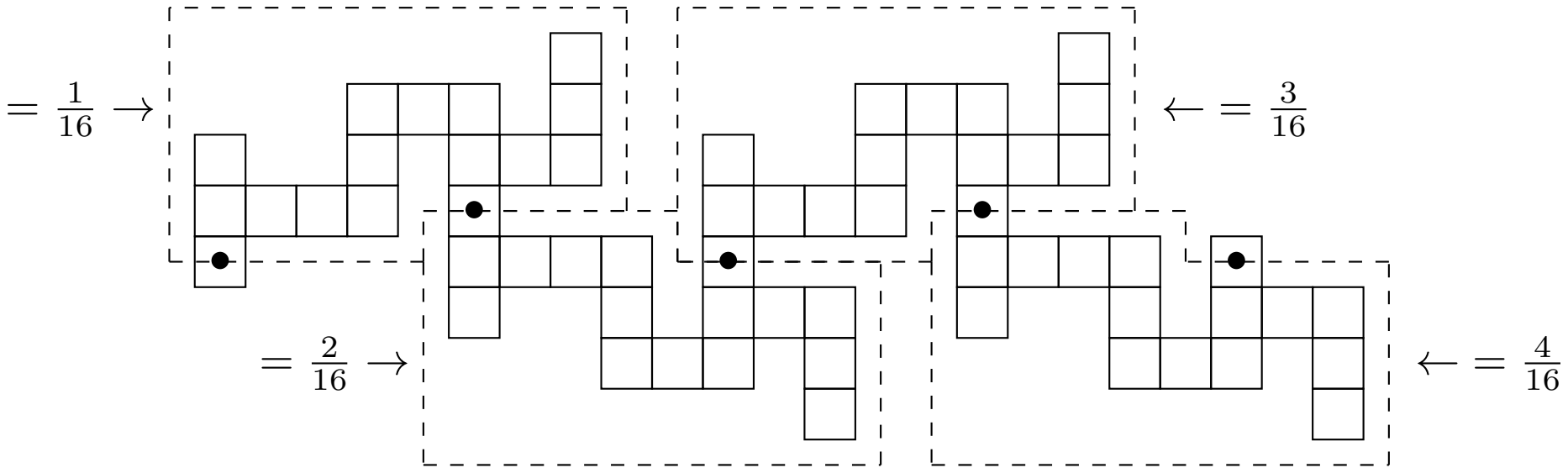

Figure 12: Constructing $k/16$ configurations for arbitrary $k$ from 1/16 fragments. Bridging squares are indicated.

Here every 1/16 fragment is linked tail-to-head with the next 1/16 fragment, where all even-numbered fragments are mirrored north-south compared with the odd-numbered fragments. Together the linked fragments form some kind of horizontal "bone" with alternating "sidewings". In this way we can build a chain of an arbitrary number $k$ of 1/16 fragments linked via explosive bridges, with value $k/16$, for any $k > 0$.

Next we have a closer look at the fragment with value 1/32 from Figure 10, as depicted in Figure 13.

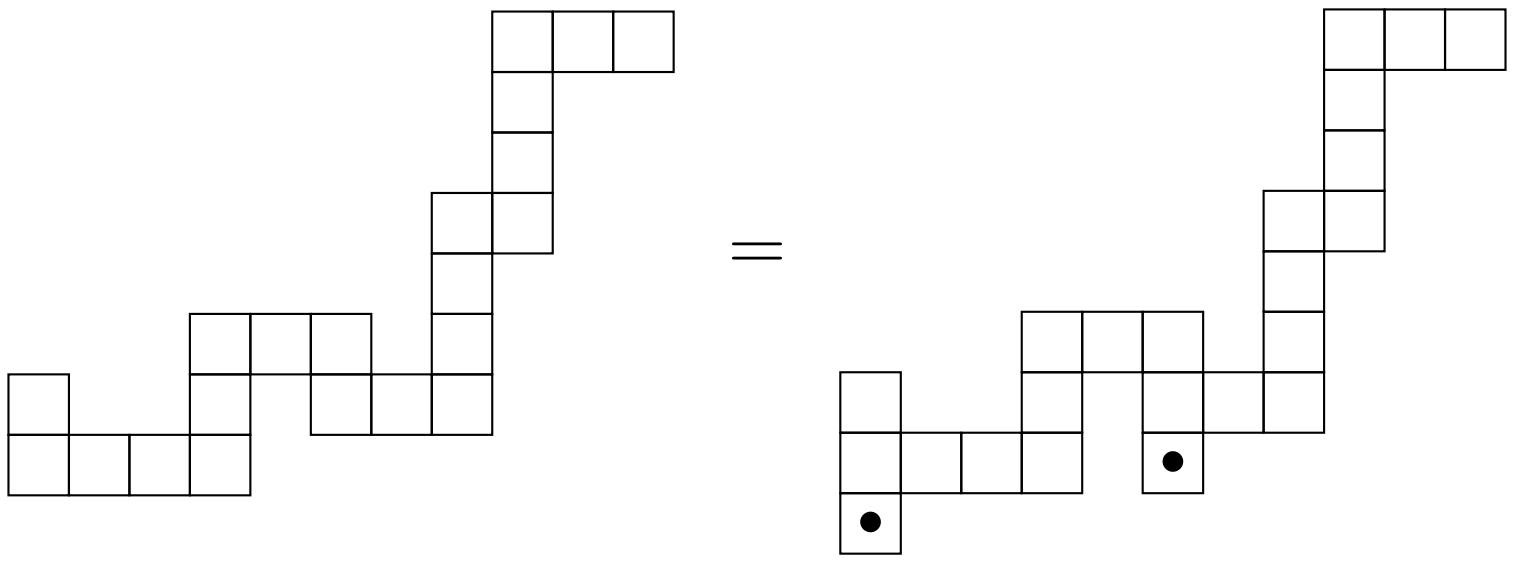

Figure 13: A fragment with value 1/32 and two explosive squares added.

We again have added the same two explosive squares as we did for the 1/16 case. As a result we have a single-component position with value 1/32, where we can use the two explosive squares as bridges with other equal fragments of value 1/32, as follows (see Figure 14).

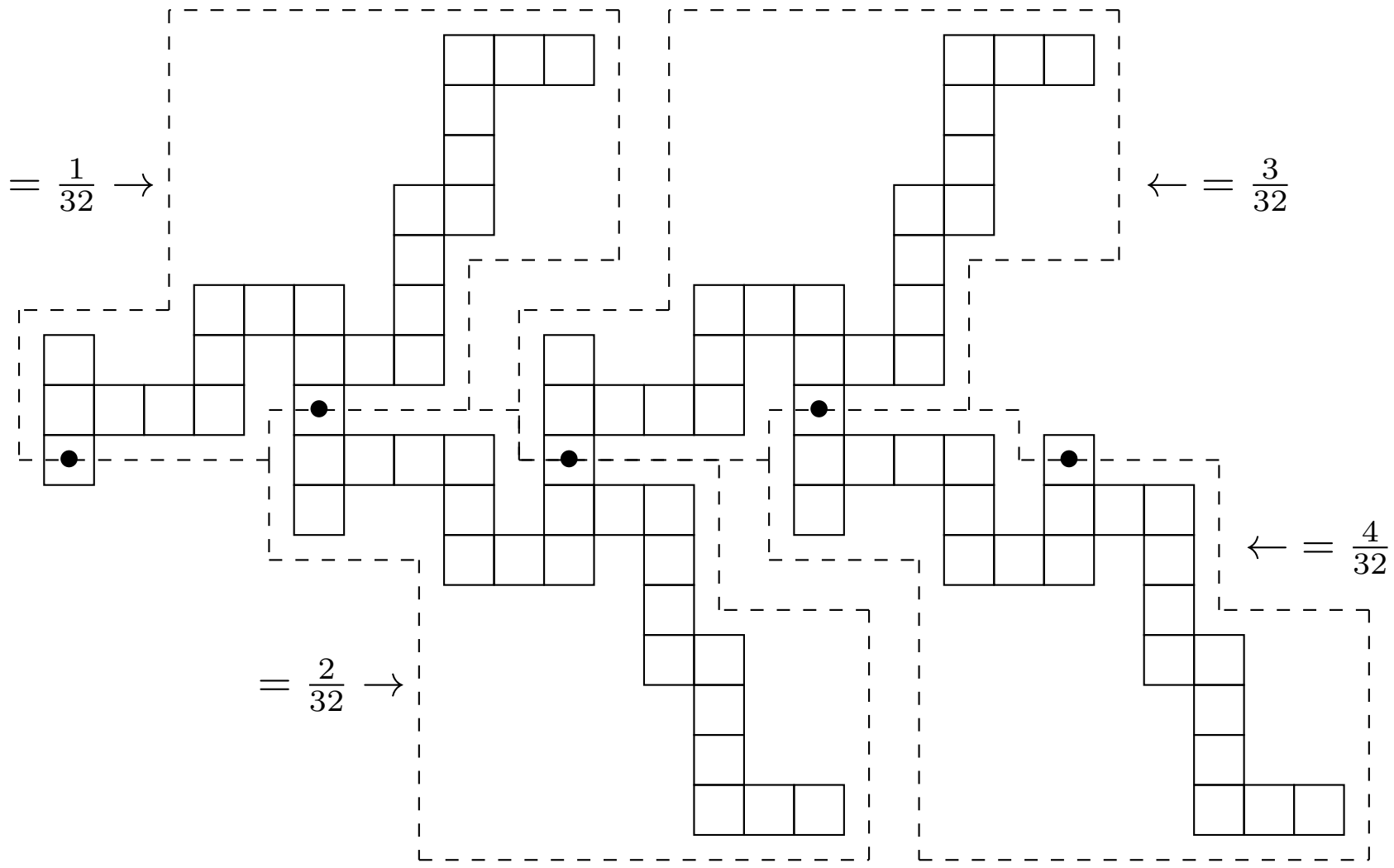

Figure 14: Constructing $k/32$ configurations for arbitrary $k$ from 1/32 fragments. Bridging squares are indicated.

We see that the chain of linked 1/32-fragments forms the same bone as the 1/16 case, but that the sidewings are elongated.

The fragments for $\frac{1}{2^n}$ for $n > 5$ as taken from Figure 10 can be used in completely similar way, with the same central bone and appropriate sidewings. A quick inspection also shows that the sidewings never will interfere with the central bone or with other sidewings. Therefore, we can build single-component positions of arbitrary multiples of $\frac{1}{2^n}$ for $n \geq 4$. This leads us to the following theorem.

**Theorem 3** (Number Theorem for Domineering)**.**
*For every CGT number for a finite combinatorial game there exists an equivalent single-component Domineering position with that number as value.*

*Proof.* Since the only possible CGT values are dyadic rationals, i.e., numbers of the form $\frac{k}{2^n}$ for $n \geq 0$ and arbitrary $k$, we only have to show that for each dyadic rational we have a single-component Domineering position with that rational as value. The proof is by construction and given in three cases.

$n = 0$: this case concerns arbitrary integers, which can be constructed as linear vertical ($\geq 0$) or horizontal ($\leq 0$) chains of length $2k$ (or $2k + 1$).

$n = 1$–$3$: these cases concern all possible multiples of 1/2, 1/4, and 1/8. For the positive fractions with values $< 1$ we have given single-component positions in Figures 8 and 9. For fractions with values $> 1$ we can add vertical fragments of even length linked upwards to the explosive squares in the fragments in Figures 8 and 9, where each two additional squares add 1 to the value, easily to verify using the Bridge Destroying Theorem. Negatives are obtained by rotating the positive positions by 90°.

$n \geq 4$: for arbitrary rationals of this type we can construct chains consisting of a central bone with sidewings, with appropriate choices of base fragments, as illustrated in Figures 12 and 14 for multiples of 1/16 and 1/32, respectively, and for any larger denominator similarly. Again, negatives are obtained by rotating the positive positions by 90°. □

*3.3. Ups and Downs*

Another interesting type of single-component positions are "pure" ups and multiple ups, e.g., $\uparrow$, $\Uparrow$, $\Uparrow\!\!\!\uparrow$, etc. We only investigate positives, since every positive is accompanied by the equivalent negative ($\downarrow$, $\Downarrow$, $\Downarrow\!\!\!\downarrow$, etc) obtained by a rotation of 90°. In our databases up to size 15 there are 15,145 different $\uparrow$ positions, the smallest being the size-6 position shown in Figure 15 left, with canonical form $\{0|*\}$. Note that there are many other single-component positions with value $\uparrow$, but they need at least a size of 7. Since they further are not of particular interest, we refrain from giving examples.

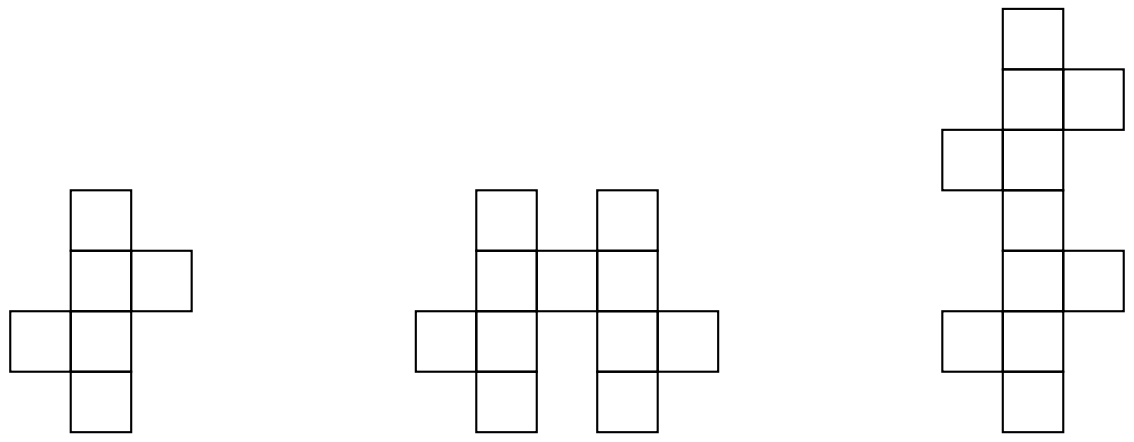

Figure 15: Up positions; left: an $\uparrow$ position of size 6; middle and right: two $\Uparrow$ positions of size 11.

In total we found 540 different positions with value $2{\cdot}\uparrow = \Uparrow$, the smallest ones of size 11 (depicted in Figure 15, middle and right). These can be understood as two $\uparrow$ positions with a 1-square overlap, connected either horizontally (middle) or vertically (right). It is easy to verify that optimal moves for Vertical always lead to an $\uparrow$. For Horizontal optimal moves lead to positions with value $\uparrow*$, either as one game, or a sum of two disjoint games. So, $\Uparrow = \{\uparrow \mid \uparrow*\}$. Using the fact that Left's move to $\uparrow$ reverses through $*$ to 0 gives the canonical form of $\Uparrow = \{0 \mid \uparrow*\}$. Note that always many equivalent positions are obtainable by reflections of parts of the positions along connecting lines. Again, the databases contain many other $\Uparrow$ positions, but they always have a size of at least 12.

The databases gave no $3{\cdot}\uparrow = \Uparrow\!\!\!\uparrow$ or larger multiples of $\uparrow$. However, it is easy to construct such single-component positions "by hand". We can build arbitrarily long chains of vertically connected $\uparrow$s. For $\Uparrow\!\!\!\uparrow$ we need 16 squares (just out of our maximum database size), for $\Uparrow\!\!\!\Uparrow = 4 \cdot \uparrow$ we need 21 squares, etc. (see Figure 16).

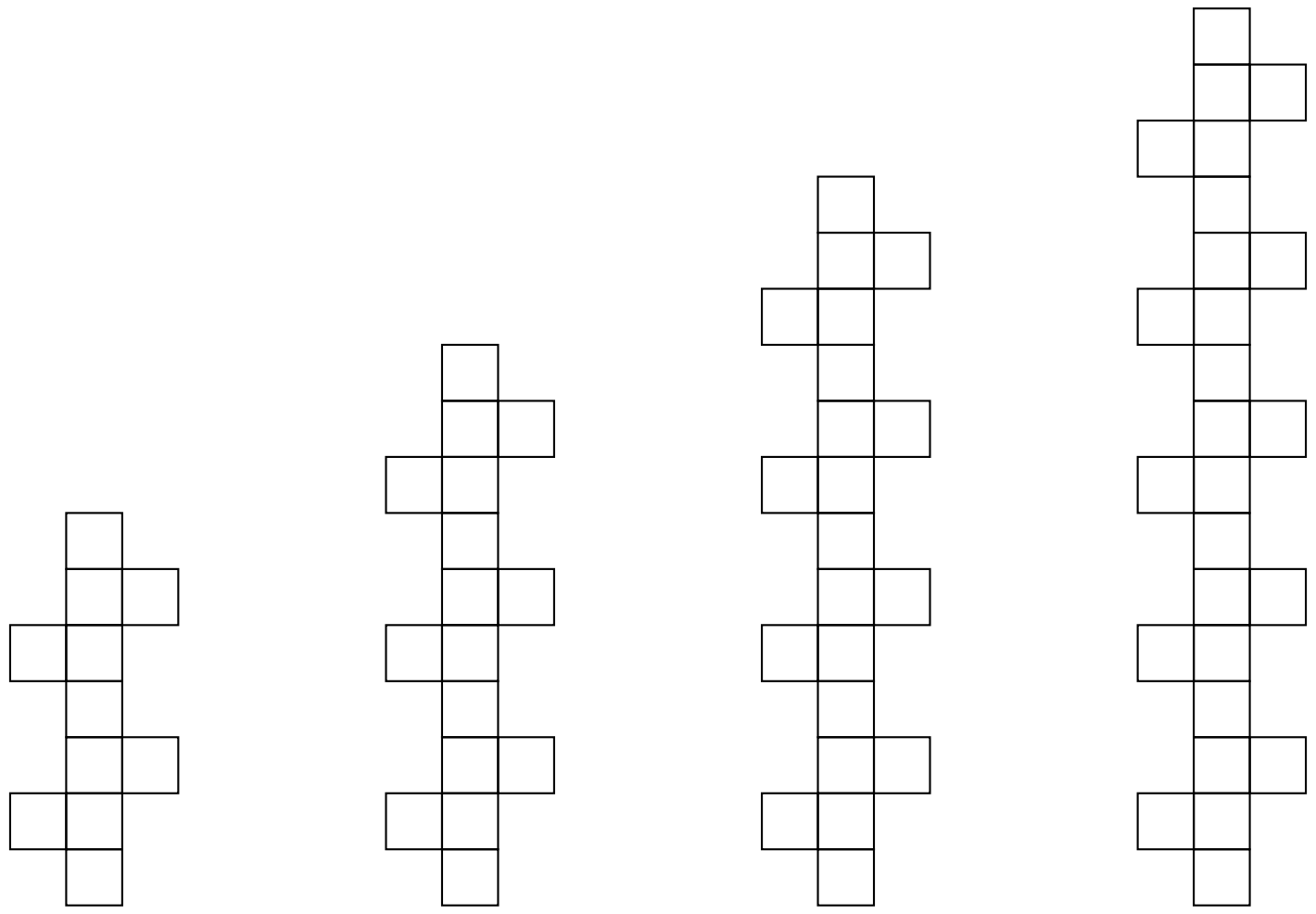

Figure 16: Constructing chains of multiple ups; from left to right: a ⇑; a ⤊; a ⟰, a $5 \cdot \uparrow$.

In this way we can continue to build $n \cdot \uparrow$ positions of size $1 + 5n$. This leads us to the following theorem.

**Theorem 4** (Ups and Downs Theorem for Domineering)**.**
*For every $n{\cdot}\uparrow$ and $n{\cdot}\downarrow$ value ($n > 0$) there exists a single-component Domineering position with that up or down as value.*

*Proof.* The proof is by construction, as shown above for arbitrary $n \cdot \uparrow$. For arbitrary $n \cdot \downarrow$ we obtain a position with that value by a rotation of 90° of a position with value $n \cdot \uparrow$. □

*3.4. Tinies and Minies*

A next interesting type of positions are "pure" tinies, of the form $⧾_x = \{0 \parallel 0 \mid -x\}$, with $x > 0$. Again, we only investigate positive positions, since every tiny is accompanied by the equivalent miny defined as $⧿_x = \{x \mid 0 \parallel 0\}$ obtained by a rotation of 90°.

In Table 2 we provide an overview of the frequencies of the different types of tinies encountered in our databases. Figure 17 shows an example with smallest size for each of these 11 categories. We omitted the trivial $⧾_0$, which is by definition $\{0 \parallel 0 \mid -0\} = \{0 \parallel 0 \mid 0\} = \{0 \mid *\} = \uparrow$. In our endgame databases we observed no multiple tinies of the form $n \cdot ⧾_x$ for $n > 0$ and arbitrary $x \neq 1$.

| $⧾_2$ | $⧾_{2*}$ | $⧾_{3/2}$ | $⧾_{3/2*}$ | $⧾_1$ | $⧾_{1*}$ | $⧾_{3/4}$ | $⧾_{1/2}$ | $⧾_{1/2*}$ | $⧾_{1/4}$ | $⧾_{1/4*}$ | total |
|---|---|---|---|---|---|---|---|---|---|---|---|
| 1037 | 4 | 11 | 3 | 25 | 19 | 29 | 231 | 115 | 39 | 8 | 1521 |

Table 2: Frequencies of $⧾_x$ positions in Domineering databases up to size 15.

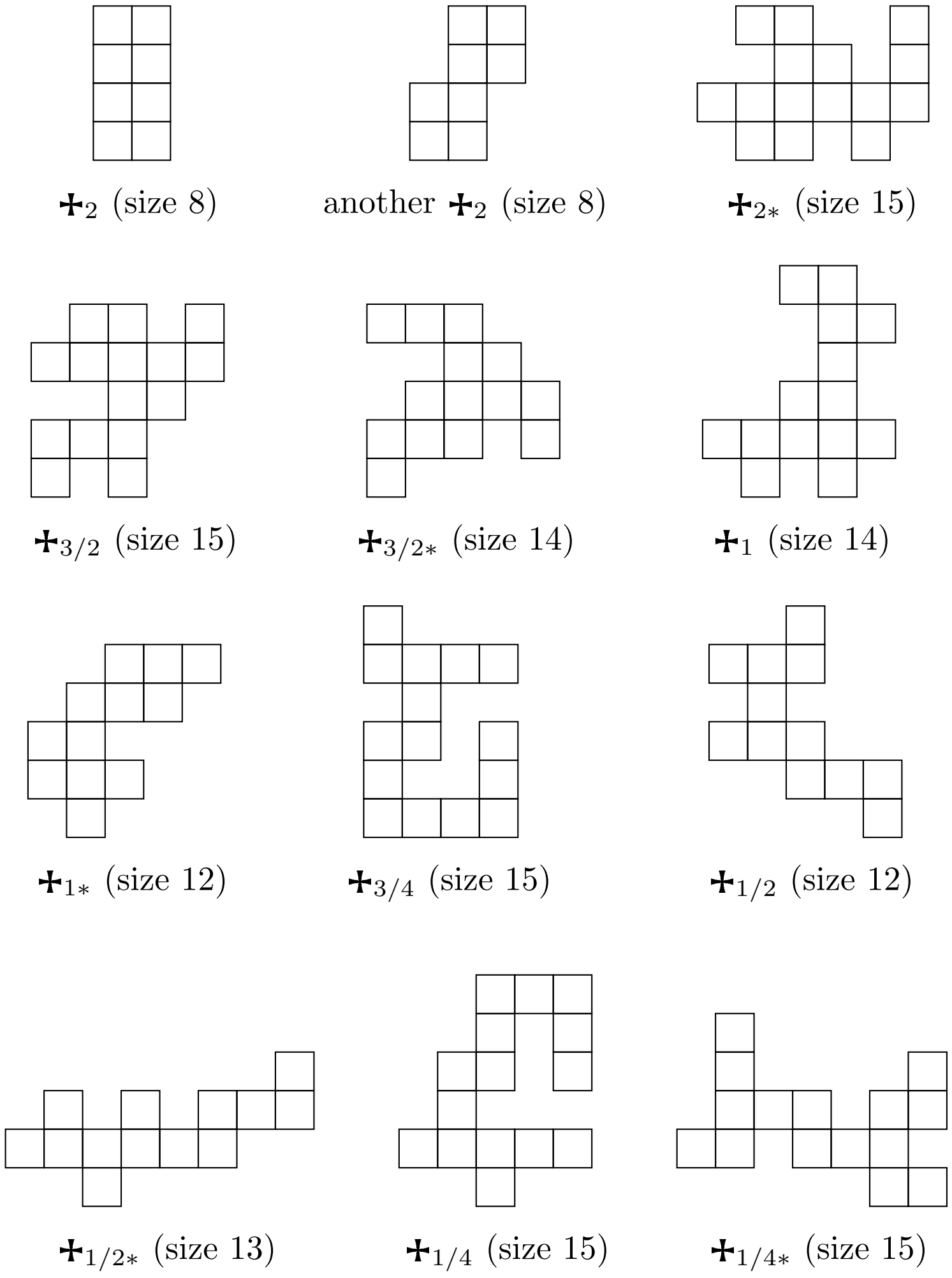

Figure 17: The two different $\boldsymbol{+}_2$'s of size 8 and examples of all other 10 tinies categories found, each for smallest database size.

### *3.5. Nimbers*

Interestingly, though simple $*$ positions are abundant in Domineering, so far no reports have been published on nimbers with higher values. Drummond-Cole states that "I have not been able to find any ordinary Domineering positions with value $*2$, and suspect that if such exist, they will be unlike the positions constructed here, which appear to depend strongly for their nimber value on the 'holes' in the corners." [10]. While he presents several positions of value $*2$, none of these can be reproduced in a game of Domineering using a regular game board. By contrast, the positions shown in Figure 18 came up in the database generation, and were found to be obtainable by normal play, i.e., in standard Domineering. They have CGT values of $*2$ and $*3$, respectively.

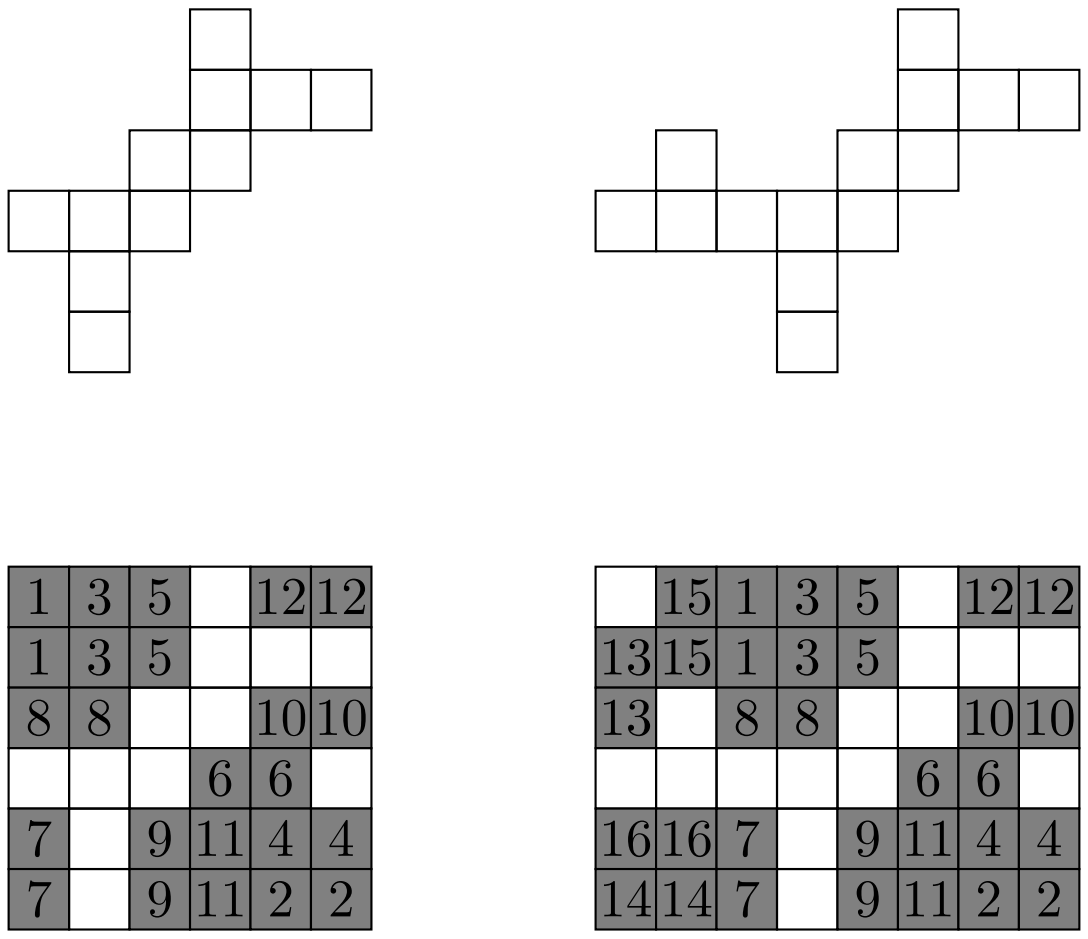

Figure 18: Two Domineering positions with values $*2$ (configuration $A$, left) and $*3$ (right). Below, the two game sequences verify that the positions can be obtained in standard Domineering.

Overall, the databases up to size 15 contain 28 distinct positions with values $*2$ and $*3$, 1 position of size 11 with value $*2$, 15 positions of size 15 with value $*2$, 8 positions of size 14 with value $*3$, and 4 positions of size 15 with value $*3$.

The left position in Figure 18 is the smallest instance of such a nimber other than $*0$ and $*1$, the one with value $*2$ of size 11. We denote this position further as configuration $A$. The eight $*3$ positions of size 14 can all be considered as consisting of the $A$ configuration with the basic $* = *1$ game of Figure 4 (the "corner" piece) attached, one such $*3$ position being shown in Figure 18 right. Since the value of the corner piece does not change when we add a third square to the central one, the Bridge Splitting Theorem shows that the value of this position is the proper sum of $*1$ and $*2$, which according to Bouton's famous analysis of the Nim game [5] equals $*3$ (the Nim addition rule). This is demonstrated in Figure 19.

There are exactly four squares in configuration $A$ where we can attach a corner piece at its central square without further interference (the squares marked with a triangle in Figure 19), and we can mirror an attached corner piece along the connection line, yielding the eight different $*3$ positions of size 14 constructed from configuration $A$. Further, since we can extend the corner piece with a third square attached to the central one (giving a small "T"-like piece) still with value $*1$, we also can attach the small-$T$ piece to one of the marked squares of position $A$, each in one orientation, which accounts for the four $*3$ positions of size 15, as illustrated in Figure 20.

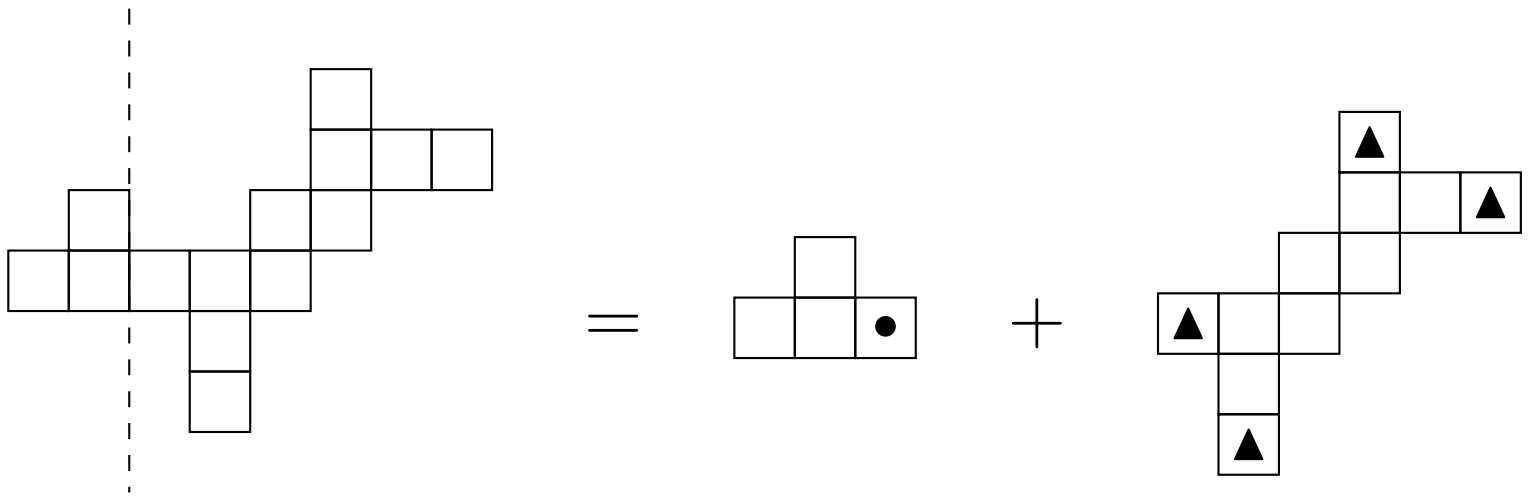

Figure 19: Showing that the left position of size 14 has value $*3 = *1 + *2$. The bullet marks the explosive square. The triangles mark the squares where the two fragments can join.

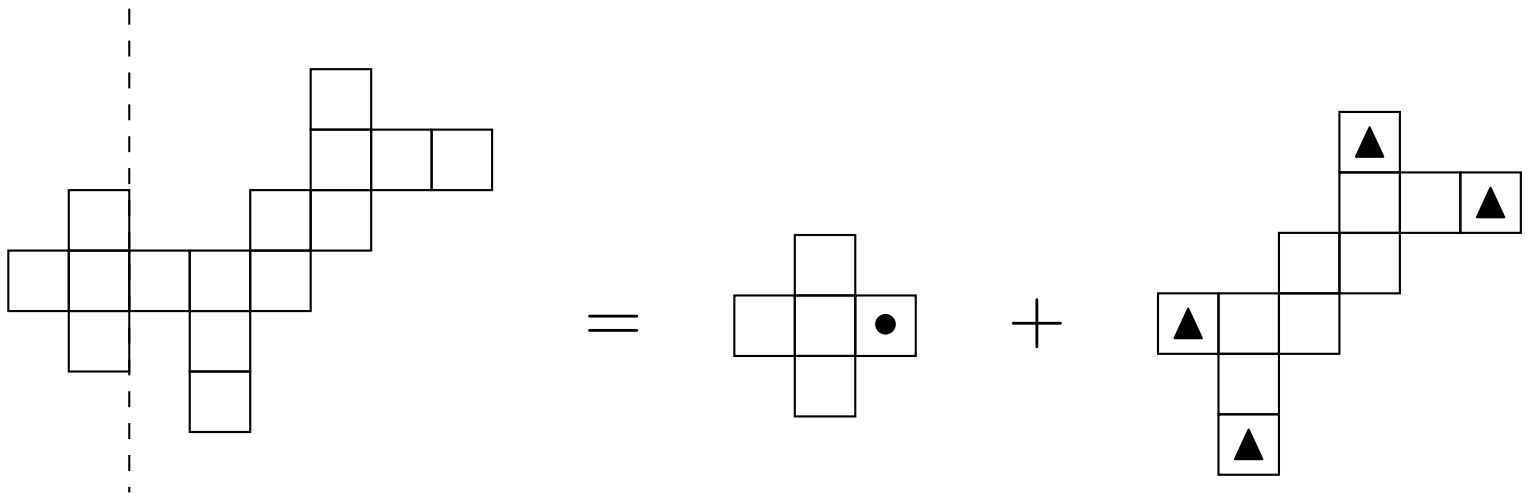

Figure 20: Showing that the left position of size 15 has value $*3 = *1 + *2$. The bullet marks the explosive square. The triangles mark the squares where the two fragments can join.

The above reasoning shows that due to the Bridge Splitting Theorem and the Nim addition rule it is easy to construct single-component $*3$ positions once we have single-component $*1$ and $*2$ positions. We can extend this further by adding a second $*1$ fragment (either the corner or the small-$T$ piece) to any of the twelve $*3$ positions built from position $A$. Since $*1 + *3 = *2$, this of course yields new positions with value $*2$. Adding a third $*1$ fragment gives again new positions with value $*3$, and finally adding a fourth $*1$ fragment gives positions of value $*2$ again, of which the largest position, with four small-$T$ pieces attached, has size 27, just fitting on a $10 \times 10$ board (see Figure 21).

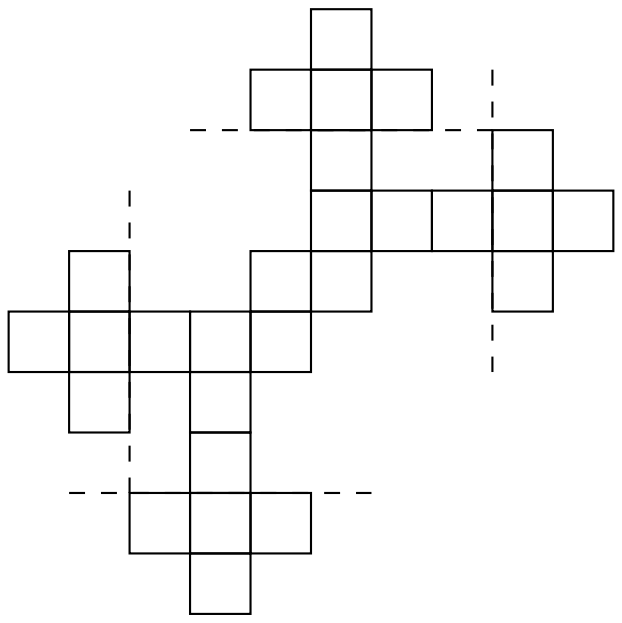

Figure 21: An example $*2$ position of size 27. The dashed lines show that this position can be considered as the sum of configuration $A$ with four small-$T$ pieces attached.

Finally, in our databases up to size 15 there are two other groups of $*2$ positions. An example of the first group, denoted as configuration $B$, is given in Figure 22 left.

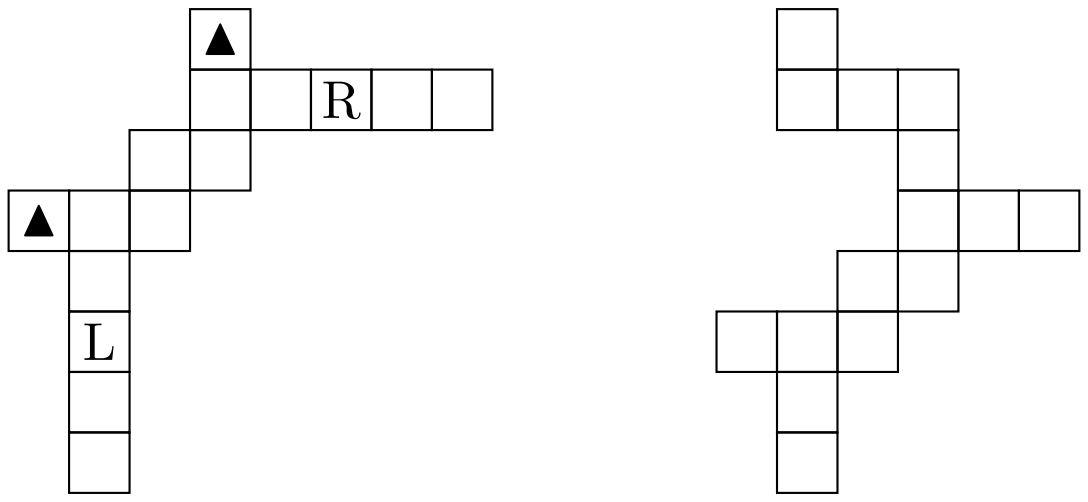

Figure 22: Two Domineering positions with value $*2$ of size 15; left: configuration $B$, right: configuration $C$.

Configuration $B$ can be derived from configuration $A$ by adding a vertical strip of length 2 to the square marked L, raising the value with $+1$, and at the same time adding a horizontal strip of length 2 to the square marked R, reducing the value with $-1$. As a result, the value is unchanged ($*2$) compared to configuration $A$. Instead of attaching to L, we could have added the vertical strip (upward) on any of the two squares marked with a triangle. Likewise, we could have used these same two marked squares also to add the horizontal strip (to the left), even simultaneously with the vertical added strip. So, in total this gives nine possibilities to add a vertical plus a horizontal strip of length 2 to configuration $A$, e.g., nine positions of configuration $B$ with value $*2$ of size 15.

Of course we can build arbitrarily large positions with value $*2$ in this way, by adding more than one vertical strip of length 2 or vertical strips of length $4, 6, 8, ...$ as long as we simultaneously add horizontal even-length strips with the same total length, and the strips do not intersect. Further, it is always possible to add an odd number of corner or small-$T$ pieces at appropriate squares, yielding arbitrarily large $*3$ positions. This proves that single-component positions with $*2$ and $*3$ (and of course the trivial $*1$ and $*0$) values are not bounded by a maximum size of the positions.

The last group of $*2$ positions of size 15 is quite different, and illustrated by the position in Figure 22 right. We denote this one as configuration $C$. This position can be seen as configuration $A$ with a negative $L$ attached to the top square of $A$, and this negative $L$ can be oriented in three ways, left-up as in Figure 22, left-down, or right-up. This accounts for three "equivalent" $C$-positions. The value is identical due to symmetry if we instead attach a positive $L$ to the left-most square of position $A$, again with three orientations, up-left, up-right, and down-left. Together this yields a group of six C-positions with value $*2$. Note that we could have attached both a negative $L$ to the top square and a positive $L$ to the left square, both in three orientations, leading to a group of nine positions with value $*2$ of size 19.

*3.6. Other Interesting Values*

In this subsection we report on positions with interesting values for one or another reason.[1]

First, we looked for positions with the largest number of (Left plus Right) canonical options. We encountered seven positions with maximum number of options, in all cases five Left plus five Right options. They all have size 15. There is just one such position fitting on a $5 \times 5$ board, given in Figure 23(a).

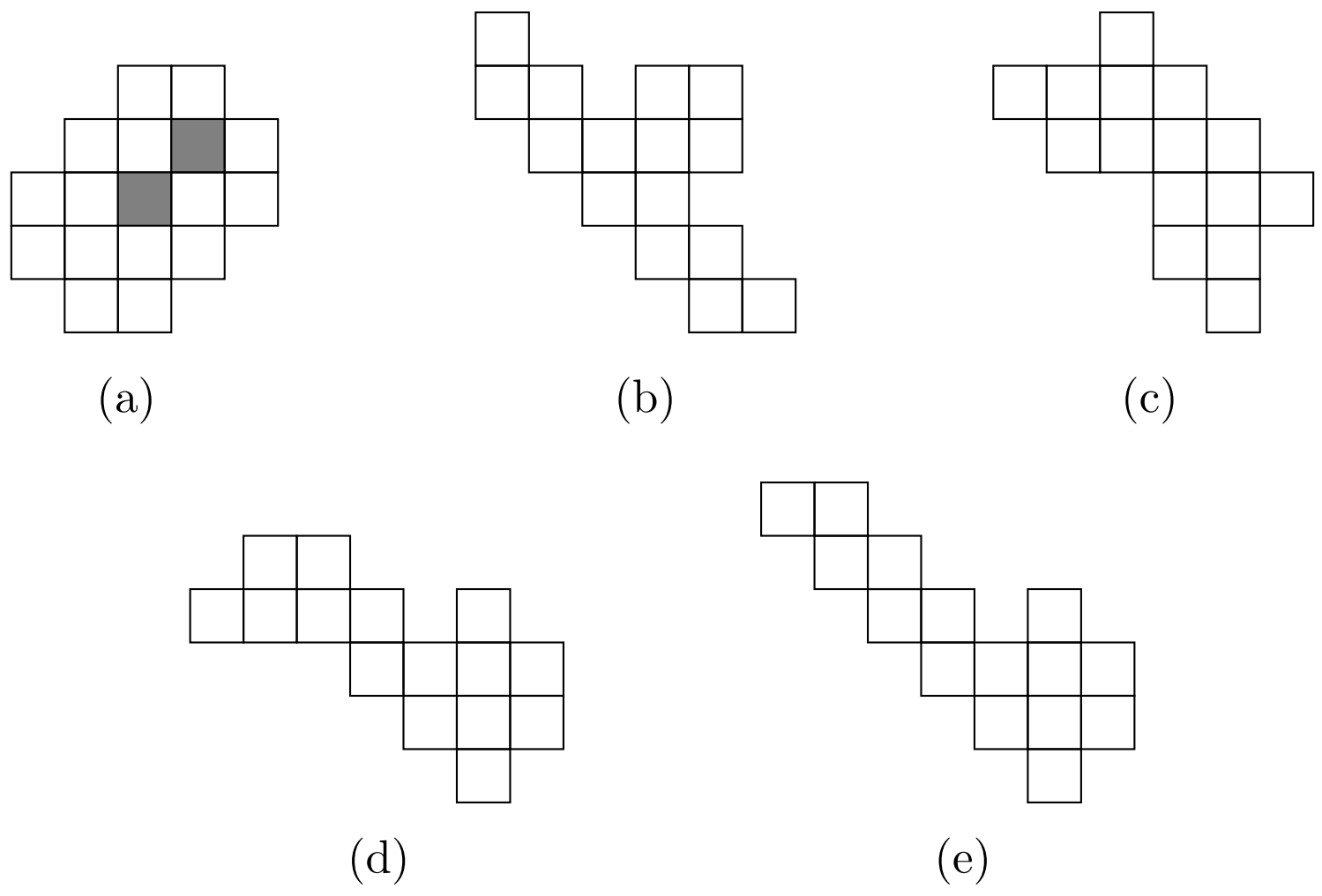

Figure 23: Domineering positions with five Left and five Right canonical options.

The grey cells in this figure denote covered squares. We immediately note that due to these inside covered squares this position is not reachable in standard Domineering, i.e., starting from a rectangular board. Drummond-Cole denotes this as a "Generalized Domineering" position [10]. Two other positions with maximum number of options fit on a $6 \times 6$ board and are given in Figures 23(b) and 23(c). Both are rotation-symmetric. Two more such positions fit on a $5 \times 7$ and $6 \times 7$ board, respectively, and are given in Figures 23(d) and 23(e). These two are not rotation-symmetric, the rotated positions accounting for the remaining two positions. All these positions except the one depicted in Figure 23(a) are reachable from appropriate rectangular boards and thus in standard Domineering.

We remark that having ten Left plus Right options constitutes no record, since Shankar and Sridharan [18] gave a position (of size 20) with even fifteen canonical options (eight Left and seven Right).

We also looked for a position with largest total number of Left plus Right canonical options, including all nested options, until only {} is left. We only found two such positions, being rotation-symmetrics, with a total of 499 Left plus Right options. One of them is depicted in Figure 24. It also is reachable in standard Domineering.

[1]For space reasons we refrain from giving the CGT values for the positions in this subsection. They can be obtained from the corresponding author on request.

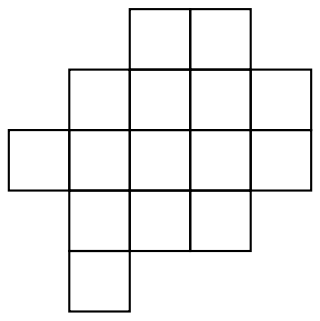

Figure 24: A Domineering position with maximum total number of Left plus Right canonical options.

Finally, just as a curiosity, we also looked for the position with largest string representation of the canonical form. We found 4 positions with largest string length, namely 475. One of them is shown in Figure 25.

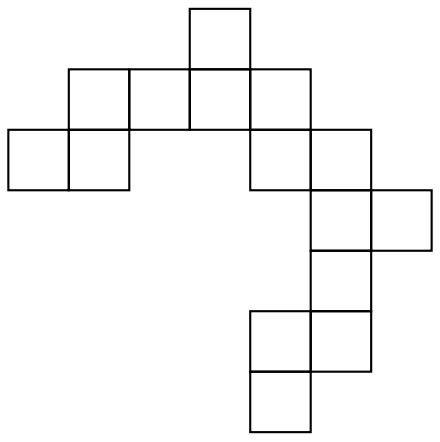

Figure 25: A Domineering position with a canonical form consisting of 475 characters.

The other three positions can easily be obtained by flipping the left corner piece along the horizontal connection line and/or the lower corner piece along the vertical connection line.

Of course a position with a string representation of the canonical form consisting of 475 characters is a far cry from the position (of size 24) shown by Shankar and Sridharan [18], whose canonical form has more than 20,000 characters. Clearly, like is the case for positions with a maximum number of (total) Left plus Right options, these numbers grow rapidly with position size, and obviously will not have upper bounds.

## 4. Temperatures in Domineering

Berlekamp conjectured that 2 is the highest possible temperature in Domineering (as reported in [18]). Drummond-Cole [9] discovered such a position (see Figure 26).

Shankar and Sridharan [18] extensively surveyed all Domineering positions fitting on boards of dimensions $5 \times 6$, $4 \times 8$, $3 \times 10$, and $2 \times 16$, searching for new temperatures. They also found the same temperature-2 position and no positions with higher temperature. In our analysis of the CGT endgame databases, as a sidetrack we also recorded the temperatures of the positions encountered. Although Shankar and Sridharan investigated many more positions, our databases were not restricted to fit within a specified area, only to have a maximum size, so it was worthwhile to look for other temperatures. However, we also found exactly one position with temperature 2, the one in Figure 26, and no positions with higher temperatures. So Berlekamp's conjecture still stands.

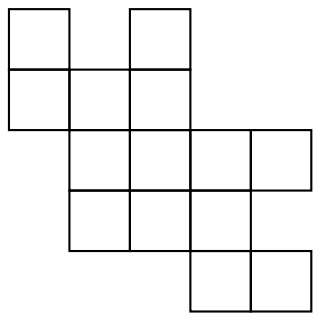

Figure 26: A Domineering position with temperature 2.

## 5. Conclusions and Future Research

We can divide our main findings in two categories. The first are of a more practical nature. Inspecting our CGT endgame databases we have discovered a large variety of interesting CGT values reachable in standard Domineering. These include many numbers, ups and downs, tinies and minies, and nimbers. In the latter category the discovery of $*2$ and $*3$ positions is of particular interest, not being found before. Moreover, we have shown that within Domineering positions up to size 15 there is exactly one (ignoring symmetrics) having temperature 2, and none with higher temperature. All these findings show that Domineering is an intriguing game, where all the pitfalls and many intricacies are yet being and still to be discovered.

Second, more theoretically, based on our findings we were able to formulate several theorems generally applicable to Domineering positions. These include the Bridge Destroying Theorem useful for (de)composing single-component positions from/into smaller fragments with known values, the Number Theorem proving that any number game in finite combinatorial games has an equivalent single-component Domineering position, and the Ups and Downs Theorem showing how single-component Domineering positions for arbitrary multiples of ups and downs can be generated.

There are several directions for future research. Along the practical lines, of course with proper refinements it should be possible to build endgame databases for even larger Domineering positions, although the maximum size reachable will be strongly restricted by memory demands. Also, as was done by Shangkar and Sridharan [18] in their search for temperatures in Domineering, larger Domineering positions can be examined without storing all results.

More theoretically, current research also approaches Domineering from a different perspective, namely using knowledge-intensive rules to solve Domineering boards without any search (we call this *perfectly solving*) [24]. It is interesting to investigate whether a combination of the CGT approach and the knowledge-intensive approach leads to more insights into Domineering.

Finally, we suggest as future research to address the following two challenges.

**Challenge 1.** *Find a Domineering position with value* $*4$.

If successful, this probably also will lead to finding single-component Domineering positions with values $*5$, $*6$, and $*7$, by combining the $*4$ position with positions with values $*1$, $*2$, and $*3$ based on the Bridge Splitting and/or Destroying Theorems, using the Nim addition rule.

**Challenge 2.** *Prove that the highest temperature possible in Domineering is 2.*

This is Berlekamp's conjecture, that is supported by much evidence by the present and other research, but not yet proven formally.